\documentclass[letterpaper]{article} 
\usepackage{aaai24}  
\usepackage{times}  
\usepackage{helvet}  
\usepackage{courier}  
\usepackage[hyphens]{url}  
\usepackage{graphicx} 
\usepackage{natbib}  
\usepackage{caption} 

\usepackage{graphicx}
\usepackage{amsmath}
\usepackage{amssymb}
\usepackage{subcaption}
\usepackage{booktabs}
\usepackage{multirow}
\usepackage{arydshln}
\usepackage[font={small}]{caption}

\usepackage{algorithm}
\usepackage{algorithmic}

\usepackage{newfloat}
\usepackage{listings}
\DeclareCaptionStyle{ruled}{labelfont=normalfont,labelsep=colon,strut=off} 
\floatstyle{ruled}
\newfloat{listing}{tb}{lst}{}
\floatname{listing}{Listing}
\title{Filter Pruning For CNN With Enhanced \\ Linear Representation Redundancy}
\author{
    Bojue Wang\textsuperscript{\rm 1},
    Chunmei Ma\textsuperscript{\rm 1}\thanks{Corresponding author},
    Bin Liu\textsuperscript{\rm 2},
    Nianbo Liu\textsuperscript{\rm 3},
    Jinqi Zhu\textsuperscript{\rm 1}
}
\affiliations{
    \textsuperscript{\rm 1}Tianjin Normal University,
	\textsuperscript{\rm 2}Southwestern University of Finance and Economics\\
	\textsuperscript{\rm 3}University of Electronic Science and Technology of China\\

    dukewbj@163.com, mcmxhd@163.com, liubin@swufe.edu.cn, liunb@uestc.edu.cn, zhujinqi1016@163.com
}

\usepackage{bibentry}

\begin{document}

\maketitle

\begin{abstract}
Structured network pruning excels non-structured methods because they can take advantage of the thriving developed parallel computing techniques. In this paper, we propose a new structured pruning method. Firstly, to create more structured redundancy, we present a data-driven loss function term calculated from the correlation coefficient matrix of different feature maps in the same layer, named CCM-loss. This loss term can encourage the neural network to learn stronger linear representation relations between feature maps during the training from the scratch so that more homogenous parts can be removed later in pruning. CCM-loss provides us with another universal transcendental mathematical tool besides $L*$-norm regularization, which concentrates on generating zeros, to generate more redundancy but for the different genres. Furthermore, we design a matching channel selection strategy based on principal components analysis to exploit the maximum potential ability of CCM-loss. In our new strategy, we mainly focus on the consistency and integrality of the information flow in the network. Instead of empirically hard-code the retain ratio for each layer, our channel selection strategy can dynamically adjust each layer's retain ratio according to the specific circumstance of a per-trained model to push the prune ratio to the limit. Notably, on the Cifar-10 dataset, our method brings 93.64\% accuracy for pruned VGG-16 with only 1.40M parameters and 49.60M FLOPs, the pruned ratios for parameters and FLOPs are 90.6\% and 84.2\%, respectively. For ResNet-50 trained on the ImageNet dataset, our approach achieves 42.8\% and 47.3\% storage and computation reductions, respectively, with an accuracy of 76.23\%. Our code is available at https://github.com/Bojue-Wang/CCM-LRR.
\end{abstract}

\section{Introduction}

Model compression is a long-term open problem, which aims to obtain smaller and faster models while maintaining accuracy. Traditionally, model compression methods can be divided into four categories: low-rank approximation\cite{tropp2023randomized}, knowledge distillation\cite{hu2022teacher}, quantization\cite{gholami2022survey} and model pruning\cite{mi2022designing}. Among them, model pruning is the most straightforward paradigm to operate on the network itself to reduce parameters and FLOPs. Model pruning can be subdivided into structured pruning and non-structured pruning. Structured pruning executes on sub-network structures like CNN filters or an entire hidden layer, etc. Thus, the pruned network maintains its structured calculating connections and still can be accelerated by utilizing general parallel computing units like GPUs.

\begin{figure}[htbp]
\begin{center}
	\begin{subfigure}{1\linewidth}
    	\includegraphics[width=1\linewidth]{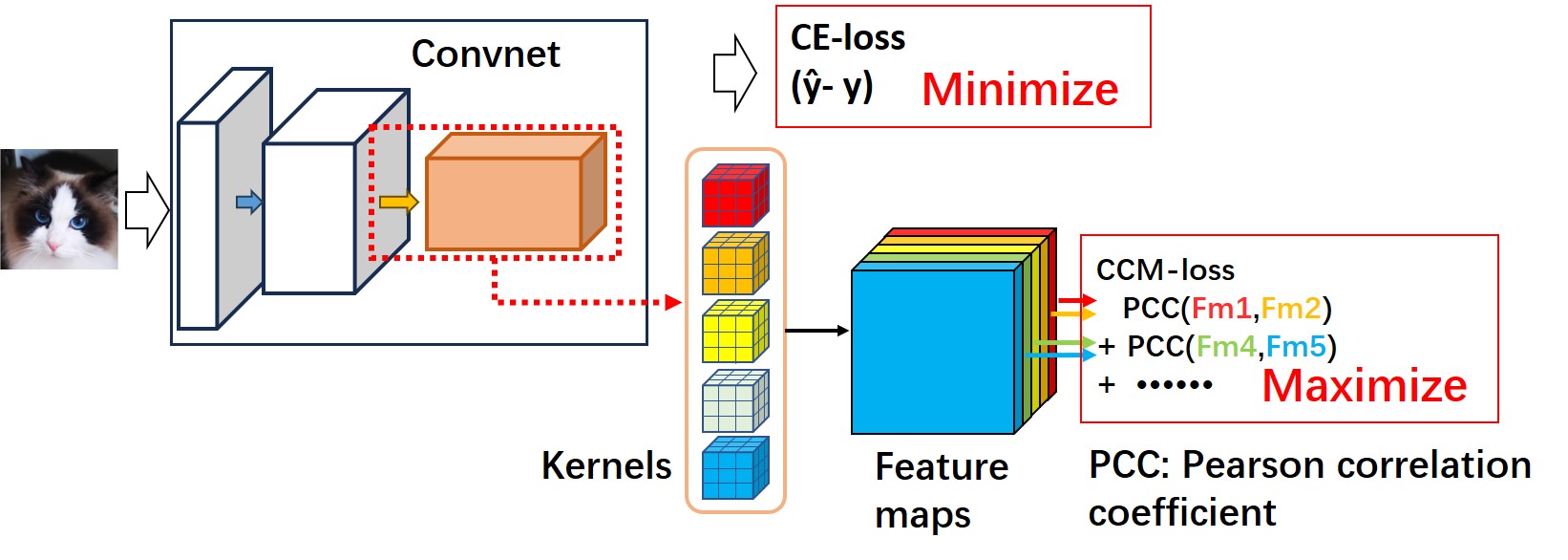}
    	\caption{Use intermediate feature maps to regulate linear property.}
    	\label{fig:short-a}
  	\end{subfigure}
	\begin{subfigure}{1\linewidth}
    	\includegraphics[width=1\linewidth]{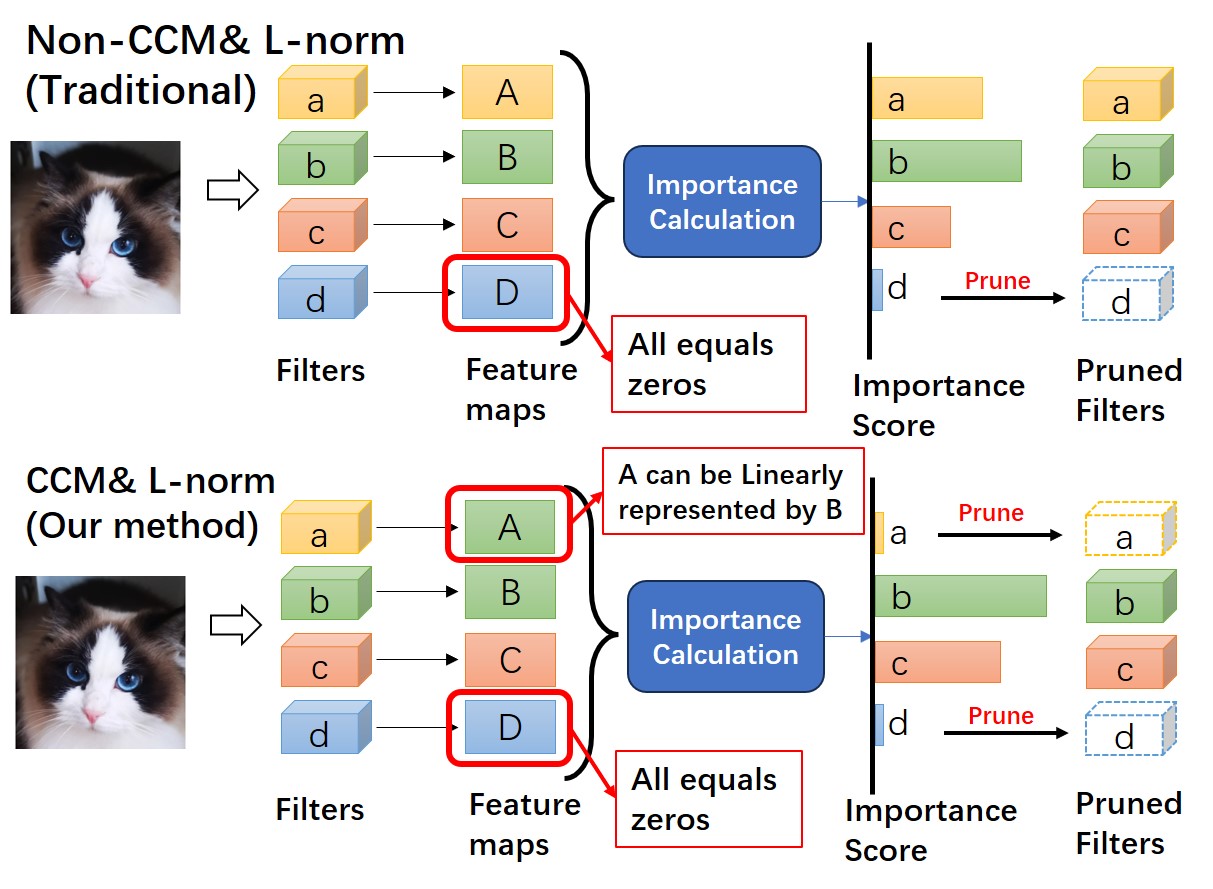}
    	\caption{Prune more filters than barely L-norm regularization.}
    	\label{fig:short-a}
  	\end{subfigure}  
\end{center}
   \caption{Our method aims to create more homogeneous feature maps during training with CCM-loss, thus we can prune more filters later in pruning.}
\label{fig:long}
\label{fig:onecol}
\end{figure}

In structured pruning tasks, creating structured redundancy and identifying which part of the parameters is redundancy are two of the most critical procedures. Traditionally, zero parameters are regarded as redundancy because they have little effect on the local or global output of the model\cite{hu2016network}\cite{li2016pruning}. L*-norm-based loss terms are the most widely used methods to generate zero redundancy. For identifying zero redundancy, both parameter-driven\cite{he2018soft} and data-driven\cite{yu2018nisp} methods have been widely studied. In recent years, researchers noted that not only zero parameters can be regarded as redundancy, but also the potential linear representation relations between sub-network structures. If some of these structures(subset A) could linearly represent others(subset B), then it can be asserted that A already carries all the information contained in B, and B can be removed. We name this phenomenon Linear Representation Redundancy (LRR). Research of the methods which tackle LRR is still in a relatively preliminary state, related works will be discussed in the section Related Works.

In this paper, we propose a new filter-level structured pruning method, shown in Fig 1. We intend to use the intermediate output from the network to regulate the network itself. Firstly, we designed a loss term to encourage the neural network to learn stronger LRR during training. We name it CCM-loss because it is calculated from the correlation coefficient matrix of the feature map channels. Maximizing CCM-loss makes the feature maps more linearly similar to each other so that more parts can be pruned in the pruning procedure. Different from L*-norm methods which concentrate on generating zeros, CCM-loss provides a new universal method to generate more redundancy. And different from other compression methods which also consider making use of linear representation property, our method needs no extra architectural designing and can be easily incorporated in existing pruning methods for better target model performance.

Next in the pruning stage, we put forward a new strategy to select channels to retain. We intend to prune the same percentage of information from each layer rather than designate a prune ratio for each layer barely on human experience. Thus, it can be prevented from pruning more information in one layer but less in others, which can cause a bottleneck of the information flow and increase the difficulty of recovery accuracy in finetuning for the model. To this end, we first measure the channel importance with CHIP\cite{sui2021chip}. CHIP is a PCA-based method, thus, a score for a channel from CHIP can be regarded as the volume of information contained in that channel. Based on this score, we calculate how much information will be pruned and where they are. Since CCM-loss compresses more information into fewer channels, the scores graded by CHIP for different filters are polarized obviously in CCM-loss settings compared with non-CCM-loss settings. Different from previous methods, the proposed strategy primarily considers reserving the consistency and integrality of the information flow to the maximum. Thus, the difficulty for the network to recover accuracy in finetuning declined. Also, different from the former methods which extensively rely on human experience to hard-code the retain ratio for each layer to designate the compression ratio of the final target model, our strategy can dynamically adjust the retain ratio for each layer and get better target model performance. Details of the proposed method will be presented in the section The Proposed Method.

We empirically apply our method in filter pruning tasks of multiple different network structures trained on the Cifar-10 and ImageNet datasets. The evaluation results show that our method excels prior works both in pruning performance and pruned model accuracy. Notably, our method brings 93.64\% accuracy with 1.4M parameters and 49.6M FLOPs for pruned VGG-16, the model size and FLOPs are reduced by 90.6\% and 84.2\%. On ResNet-56 and ResNet-110, Our experiments show in several layers, even if we left only one filter, it still can support the model to get excellent performance on accuracy. On the ImageNet dataset, our approach can achieve 76.48\% accuracy while we prune 42.8\% of the parameters and 47.3\% of FLOPs, bringing +0.02\% over our baseline model. After we chop off 78.5\% of the parameters and 79.9\% of FLOPs, our accuracy still can remain at 72.76\%. The experiment and evaluation results will be described in the section Experiments. In the section Conclusion And Discussion, we talk about the extending value of our method and propose some suggestions for future exploration.

\section{Related Works}

For a long time, zero parameters are the only kind of redundancy that model pruning aims to tackle because they have little effect on the output of the model. $L_{1}$,$L_{2}$-norm regularization are the most typical approach to generate zeros and had been widely used in nowadays neural network \cite{kumar2021pruning}\cite{wen2016learning}\cite{he2018soft}. To avoid the drawback of irregular connections, which are unfriendly to parallel computing, structured pruning methods like SSL \cite{wen2016learning} were proposed. SSL uses Group Lasso regularization \cite{yuan2006model} as its theoretical basis to generate zeros on structural grouped parameters, basically a variation of L-norm methods. \cite{li2016pruning} measures the importance of a filter by calculating the sum of its absolute weights. APoZ \cite{hu2016network} measures the importance of the filter by the percentage of zero activations of a neuron after ReLU mapping, it is a data-driven method but still aims to tackle zeros. Many works try to measure the importance of filters in different ways. For example, HRank \cite{lin2020hrank} selects the proper feature maps with the change of rank when one feature map is removed, CHIP \cite{sui2021chip} uses the change of nuclear norm instead of rank to grade each filter, etc. These methods do not explicitly tackle zeros any more, they try to grade filters with a more general approach. But still, LRR receives no special attention.

In recent years, a few works begin to lay their eyes on LRR. Centripetal SGD \cite{ding2019centripetal} use K-means clustering to shape the filters in the same cluster to be similar to each other while training from scratch, but the number of clusters, which equals the remained filters after pruning, still relies on human experience to designate. GhostNet \cite{han2020ghostnet} redesigned the network structure. Still in their method, the filter number of a layer is written, and to get enough channels to feed into the next layer, they use linear operations to generate extra feature map channels, which we believe is actually creating unnecessary channels for the next layer to process.

\begin{figure*}[htbp]
  \centering
  \includegraphics[width=0.7\linewidth]{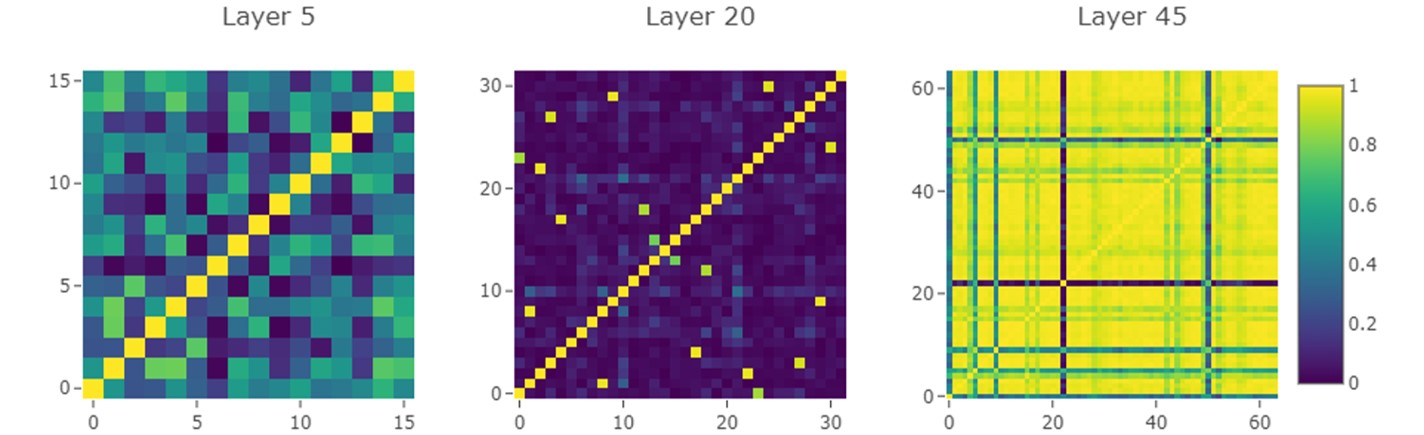}
   \caption{Examples of correlation coefficient matrixes manifested in heatmap style. The brighter a pixel is, the stronger correlation it indicates. All equal to $1$ on the diagonal, which is the correlation coefficient of a channel and itself. The three pictures correspond to the None/Plus/Minus version of CCM-loss for 16/32/64 channels respectively. Data come from ResNet-56 trained on the Cifar-10 dataset. Here we can see a sharp contrast of training effect between three different settings.
   }
   \label{fig:short}
\end{figure*}

\section{The Proposed Method}

Our method can be divided into three stages: train from scratch, channel importance evaluation, prune and finetune. In the first stage, we use CCM-loss to learn more redundancy. In the second stage, we use CHIP to grade each channel. And in the last, we propose our pruning strategy which concerns about the consistency and integrality of the information flow in the network.

\subsection{Correlation Coefficient Matrix Loss}

To encourage CNN-based neural networks to generate more Linear Representation Redundance on filter level, we design a loss term using correlation coefficients between different feature map channels.

Let 
$\mathcal{F}_{l} = \{\mathcal{C}_{l}^{1},\mathcal{C}_{l}^{2},\cdot \cdot \cdot, \mathcal{C}_{l}^{n_{l}} \}\in  \mathbb{R}^{b\times n_{l} \times h_{l}\times w_{l}}$ be the output feature map of the $l$-th layer, which
contains 
$n_{l}$
channels
$\mathcal{C}_{l}^{k} \in  \mathbb{R}^{h_{l}\times w_{l}}$ ,
 $b$ is batch-size. Firstly, we calculate the average through the batch, get 
$\bar{\mathcal{F}}_{l} = \{\bar{\mathcal{C}}_{l}^{1},\bar{\mathcal{C}}_{l}^{2},\cdot \cdot \cdot, \bar{\mathcal{C}}_{l}^{n_{l}} \}\in  \mathbb{R}^{n_{l} \times h_{l}\times w_{l}}$ . Then we flatten each channel, resize $\bar{\mathcal{C}}_{l}^{k}$ to $\mathcal{R}_{l}^{k}\in  \mathbb{R}^{1 \times h_{l} w_{l}}$, and get 
$\tilde{\mathcal{F}}_{l} = \{\mathcal{R}_{l}^{1},\mathcal{R}_{l}^{2},\cdot \cdot \cdot, \mathcal{R}_{l}^{n_{l}} \}\in  \mathbb{R}^{n_{l} \times h_{l} w_{l}}$
. The correlation coefficient of two channels is define as follow:

\begin{equation}
\small
r(\mathcal{R}_{l}^{i},\mathcal{R}_{l}^{j})=\frac{Cov(\mathcal{R}_{l}^{i},\mathcal{R}_{l}^{j})}{\sqrt{Var[\mathcal{R}_{l}^{i}]Var[\mathcal{R}_{l}^{j}]}}
\end{equation}

We calculate correlation coefficient of all channel pairs in the $l$-th layer and get the correlation coefficient matrix of the $n_{l}$ channels $\mathcal{M}_{l}\in \mathbb{R}^{n_{l} \times n_{l}}$. Consider if the absolute number of correlation coefficient between $\mathcal{Y}$ and $\mathcal{X}$ equals $1$, then $\exists a,b \:\:  s.t.\: \: \mathcal{P}\{\mathcal{Y} = a\cdot \mathcal{X}+b\} =1$,  here we take the absolute values of the correlation coefficients. After that, we sum up all the correlation coefficients in the $l$-th layer`s correlation coefficient matrix. Here we get the $i$-th layer correlation coefficient matrix loss, short by CCM-loss. For channel count in each layer may variate in a large range and the volume of correlation coefficient matrix is the square of channel count, which makes CCM-loss of different layers float dramatically, we divide $l$-th CCM-loss by the count of corresponding correlation coefficient matrix's element to fix this problem.

\begin{equation}
\small
\mathcal{L}_{CCM_{l-th}} = \frac{1}{n_{l}\times n_{l}}(\sum_{i=1}^{n_{l}}\sum _{j=1}^{n_{l}}\left |r(\mathcal{R}_{l}^{i},\mathcal{R}_{l}^{j})\right |)
\end{equation}

Sum up all the target $L$ layers` CCM-losses as the final CCM-loss, multiply with a ratio control coefficient positive number $\lambda$, and combine with the classification cross-entropy loss, and we get the final objective function: 

\begin{equation}
\small
\mathcal{L}_{obj} = \mathcal{L}_{ce}-\lambda( \sum _{l=1}^{L}\mathcal{L}_{CCM_{l-th}})
\end{equation}

\begin{figure*}[htbp]
  \centering
  \includegraphics[width=0.65\linewidth]{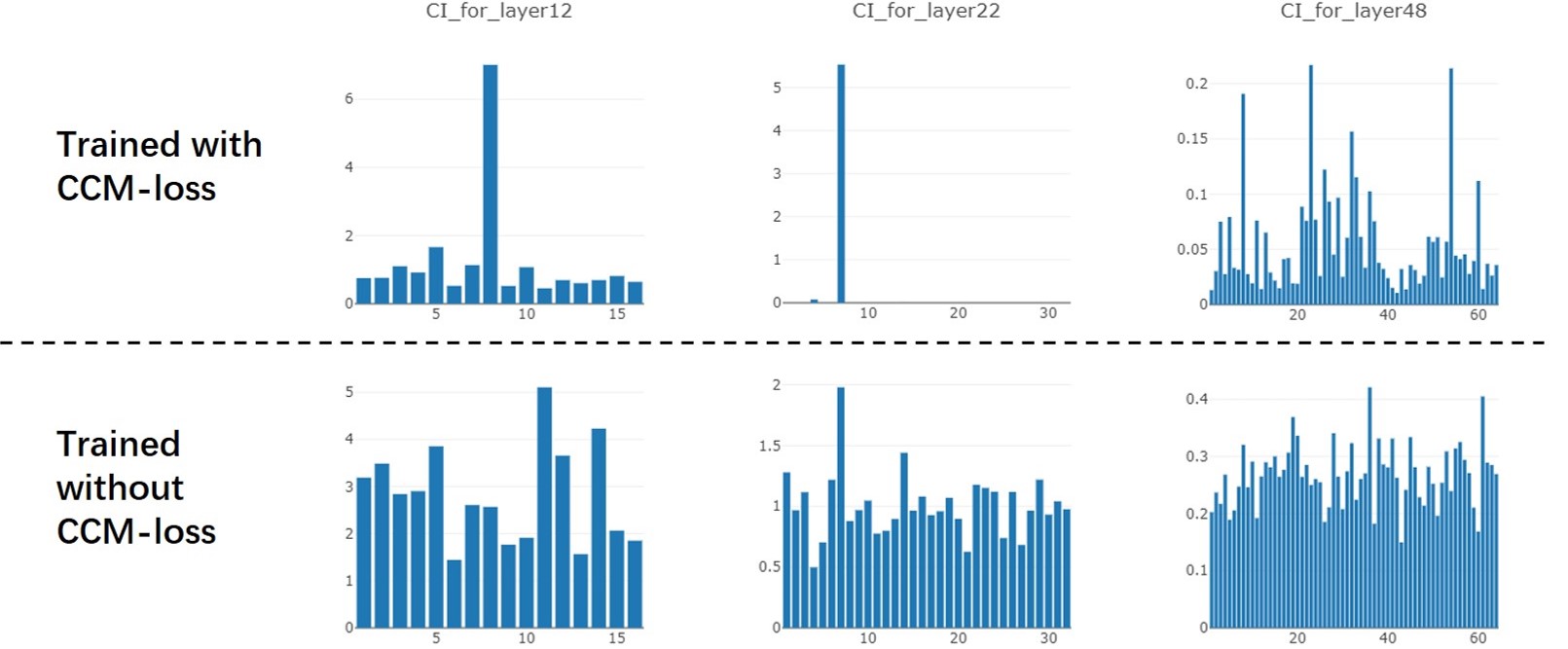}
   \caption{ Channel importances visualization for the feature map from different layers of ResNet-56. The upper three are trained with CCM-loss, and the under three are the according layers trained without CCM-loss. Compare between upper and under, recognize in CCM-loss setting, channel importances are polarized obviously.
   }
   \label{fig:onecol}

\end{figure*}

Notice here that we let the operation before CCM-loss be minus, which means while we minimize the CE loss, we want to push the correlation of all channels to be stronger. CCM-loss also can be used to push the correlation of the channels weaken from each other. To do so, we just need to change the operation to plus. In this paper, we mainly focus on the minus version. Fig.2 shows the impact of CCM-loss on the channel correlation.

\subsection{Channel Importance Evaluation}

The next step is to find out which channel subset is more important. We inherit the channel importance criterion from CHIP\cite{sui2021chip}. In the method, channel importance is calculated by the change of nuclear norm of the entire set of feature maps when masking the target channel to zeros. The nuclear norm of a matrix is the sum of the matrix's singular values, which is one of the PCA methods. CHIP measures channel importance from the inter-channel perspective and signs a quantificational score for each channel, which is necessary for the next step. For more detail about CHIP, please refer to \cite{sui2021chip}. The change of channel importances between with or without the CCM-loss term is shown in Fig.3. We can see that incorporating the CCM-loss term during training makes the importances of different channels polarize obviously.

\subsection{A New Strategy For Channel Retaining}

From Fig.3, take layer 22nd as an example, we can observe that by our evaluation criteria to measure the channel importance, most of the channels became inessential when the network was trained with the setting of CCM-loss. In prior works, the retain ratio of each layer are hard-coded in the first place, then prune the least important channels until the ratio is satisfied. This process highly depends on empirical judgment to set the retain ratio for each layer and does not considering to push the retain ratio to the limit. If we follow this paradigm, it will be easy for us to retain much more unnecessary space than we actually need. For example, in the 22nd layer with the CCM-loss setting, if we reserve 30\% of the filters, a common empirical percentage,will left us 10 from 32 original filters. From Fig.3, seemingly only one filter is essential, which makes 9 of the left 10 unnecessary. 

Here we put forward a pair of concepts: the consistency and integrality of the principal component of the information flow. The consistency of information flow is proposed to describe the information content in each layer of a well-trained network should be the same. That is because when a network is already well trained, the information content in each layer contains all the information for the network to make a certain decision, thus, the information content in each layer is equal to each other between different layers. When we prune the network, if we pruned too much information in one layer but lesser in other layers, the one layer then becomes the bottleneck for better accuracy. To reserve consistency is to keep the percentage of the retained principal component in each layer to be equal, and prevent the formation of bottlenecks while pruning. The integrality of information flow is proposed to describe how much information we tried to prune in a specific layer, the lesser we prune the better integrality we keep.

It may break the consistency and integrality if we confine the retain ratio in the first place, and a bad set of retain ratio hyper-parameter may highly increase the difficulty of the fine-tuning process to recover the accuracy. For example, in the 48th layer without CCM-loss setting, if the reserve rate is 50\%, will leave us 32 from the 64 original filters, then we lost the other 50\% filters which may carry useful information. The lost information has to be relearned during the fine-tuning process.

To tickle these problems, we design a new channel retaining strategy. After we calculate importance for each channel with the method in section 3.2, we got a list for $l$-th layer $\mathcal{L}_{l} = \{\mathcal{C}_{l}^{1},\mathcal{C}_{l}^{2},\cdot \cdot \cdot, \mathcal{C}_{l}^{n_{l}} \}$, which contains $n_{l}$ channels, $\mathcal{C}_{l}^{k}$ is the channel importance of the $k$-th channel. Firstly, we sort the list in descending order $\mathcal{S}_{l} = \{\mathcal{S}_{l}^{m_{1}},\mathcal{S}_{l}^{m_{2}},\cdot \cdot \cdot, \mathcal{S}_{l}^{m_{n_{l}}} \}$, $m_{k}$ is the index of $\mathcal{S}_{l}^{m_{k}}$ in $\mathcal{L}_{l}$ and $k$ is the index in $\mathcal{S}_{l}$. We select the first k element with the following inequations:

\begin{equation}
\small
\frac{\sum _{1}^{k-1}(\mathcal{S}_{l}^{m_{1}},\cdot \cdot \cdot ,\mathcal{S}_{l}^{m_{k-1}})}{\sum _{1}^{n_{l}}(\mathcal{C}_{l}^{1},\cdot \cdot \cdot ,\mathcal{C}_{l}^{n_{l}})}<  \alpha 
\end{equation}

\begin{equation}
\small
\frac{\sum _{1}^{k}(\mathcal{S}_{l}^{m_{1}},\cdot \cdot \cdot ,\mathcal{S}_{l}^{m_{k}})}{\sum _{1}^{n_{l}}(\mathcal{C}_{l}^{1},\cdot \cdot \cdot ,\mathcal{C}_{l}^{n_{l}})}\geq   \alpha 
\end{equation}

where $\alpha \in (0,1) $ is named \textbf{principal component retain ratio(PCRR)}. The same $\alpha $ is used in all target layers to indicate the percentage of the principal component we would like to retain in each layer. Thus, we kept the consistency and integrality of the principal component information flow. Here we obtain the remained channel set $\mathcal{RM}_{l} = \{m_{1},m_{2},\cdot \cdot \cdot, m_{k} \}$, as shown in Fig.4.

\begin{figure}[htbp]
\begin{center}
   \includegraphics[width=0.85\linewidth]{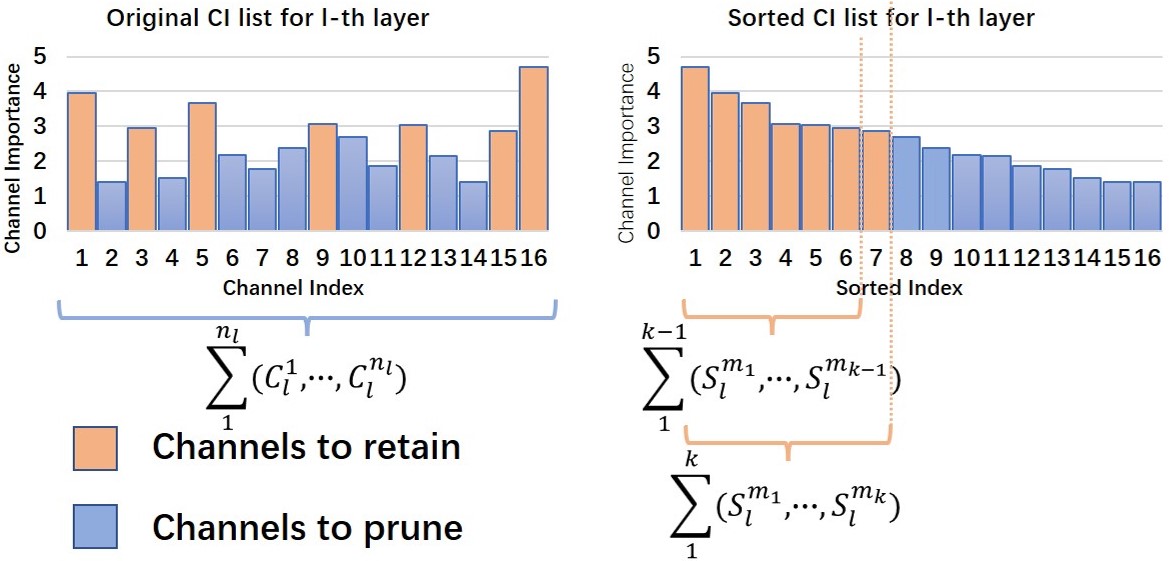}
\end{center}
   \caption{Visualization of the proposed channel selection strategy.}
\label{fig:long}
\label{fig:onecol}
\end{figure}

Considering CCM-loss helps polarize channel importances, this new strategy can collaborate with CCM-loss to select several most useful channels to retain. Rather than hard-coded for each layer empirically, for a given $\alpha$, our strategy can dynamically adjust the retain ratio for each layer according to the specific circumstance of a pre-trained model. Retrospect Fig.3, for the 22nd layer with CCM-loss, our experiments show, even if we kept only one channel, which achieved the maximum expectation of filter-level structured pruning, the accuracy of the ultimate fine-tuned model can still be maintained at a high level. Details about our experiments will be talked about in the next section.

\section{Experiments}

\subsection{Experimental Settings}
\textbf{Datasets and Baselines.} To demonstrate the effectiveness and generality of our method on filter-level structured pruning, we conduct experiments on the CIFAR-10 dataset \cite{krizhevsky2009learning} for three mainstream convolutional-based network structures: VGG-16\cite{simonyan2014very}, ResNet-56, and ResNet-110\cite{he2016deep}. We use our per-trained model as the baseline, these models have better accuracy compared with the same model that had been trained a few years ago and mentioned in previous works. We also compare our method`s performance with other state-of-art pruning methods.

\textbf{Evaluation Protocols.} We adopt the widely-used protocols, \textit{i.e.}, number of parameters and required Float Points Operations(FLOPs), to evaluate model size and computational requirement. We provide top-1 accuracy and the pruning rate of our pruned models. Also, we compare the retained filter number with per-trained models and the pruned models from CHIP to empirically analyze the compression ability of our method.


\begin{table*}[htbp]
\footnotesize
\setlength{\abovecaptionskip}{0cm}
\begin{center}
\begin{tabular}{lccccc}
\toprule
\multirow{2}{*}{Method}  & \multicolumn{3}{c}{Top-1 Accuracy(\%)} & \multirow{2}{*} {Params($\downarrow$\%)} & \multirow{2}{*}{FLOPs($\downarrow$\%)} \\
\cmidrule{2-4}
& Baseline & Pruned & $\Delta$  \\

\midrule
$l_{1}$-norm\cite{li2016pruning} & 93.96 & 93.40 & -0.56 & 5.40M(64.0) & 206.00M(34.3)\\
SSS\cite{huang2018data} & 93.96 & 93.02 & -0.94 & 3.93M(73.8) & 183.13M(41.6)\\
GAL\cite{lin2019towards}  & 93.96 & 93.77 & -0.19 & 3.36M(77.6) & 189.49M(39.6) \\
Ghost-VGG-16 (s=2)\cite{han2020ghostnet} & 93.60 & 93.70 & +0.10 & 7.70M(48.7) & 158.00M(49.5)\\
HRank\cite{lin2020hrank}  & 93.96 & 93.43 & -0.53 & 2.51M(82.9) & 145.61M(53.5) \\
CHIP\cite{sui2021chip}  & 93.96 & 93.86 & -0.10 & 2.76M(81.6) & 131.17M(58.1) \\
\hdashline
GAL\cite{lin2019towards}  & 93.96 & 93.42 & -0,54 & 2.67M(82.2) & 171.89M(45.2) \\
HRank\cite{lin2020hrank}  & 93.96 & 92.34 & -1.62 & 2.64M(82.1) & 108.61M(65.3) \\
CHIP\cite{sui2021chip}  & 93.96 & 93.72 & -0.24 & 2.50M(83.3) & 104.78M(66.6) \\
\hdashline
HRank\cite{lin2020hrank}  & 93.96 & 91.23 & -2.73 & 1.78M(92.0) & 73.70M(76.5) \\
$\star$ CHIP\cite{sui2021chip}  & 93.96 & 93.18 & -0.78 & 1.90M(87.3) & 66.95M(78.6) \\
\hdashline
Baseline(non-CCM-loss,unpruned)  & 94.27 & - & - & 14.99M(0.0) & 314.57M(0.0) \\
Baseline(CCM-loss,unpruned)  & 94.29 & - & - & - & - \\
Ours(PCRR=0.9)  & 94.29 & \textbf{94.56} & +0.27 & 4.81M(67.9) & 163.76M(47.9) \\
Ours(PCRR=0.8)  & 94.29 & \textbf{94.23} & -0.06 & 3.13M(79.1) & 110.99M(64.7) \\
Ours(PCRR=0.7)  & 94.29 & \textbf{94.12} & -0.17 & 2.09M(86.0) & 75.69M(75.9) \\
$\star$ Ours(PCRR=0.6)  & 94.29 & \textbf{93.64} & -0.65 & \textbf{1.40M(90.6)} & \textbf{49.60M(84.2)} \\

\bottomrule
\end{tabular}
\end{center}
\caption{VGG-16 on Cifar-10.}
\label{fig:long}
\label{fig:onecol}
\end{table*}

\textbf{Configurations.} We use PyTorch\cite{paszke2017automatic} to implement our method, the backbone code is based on CHIP. For both training from scratch and finetuning, we solve the optimization problem by using Stochastic Gradient Descent algorithm (SGD) with an initial learning rate of 0.01. For experiments on Cifar-10, the batch size, weight decay, and momentum are set to 256, 0.005, and 0.9, respectively. For the per-trained models with CCM-loss, the epochs are set to 400, and $\lambda$ for the CCM-loss term is set to 0.01, 0.01, and 0.004 for VGG-16, ResNet-56, and ResNet-110, respectively. The epochs in finetune proceess is set to 600. For experiments ResNet-50 trained on ImageNet, batch size equals 1920, epochs for both train from scratch and finetune is 180, and $\lambda$ equals 0.07. We use CosineAnnealingLR to adjust the learning rate during the training for all the training processes. We adopted one time shot manner, after we get the per-trained model, we conduct pruning on the whole model only one time and then go to fine-tuning. For Cifar-10, all experiments are conducted on one NVIDIA GTX 1080Ti GPU. For ImageNet, all experiments are conducted on four NVIDIA V100.

\subsection{ Results and Analysis}

\subsubsection{VGG-16 on Cifar-10}

Tab.1 shows the performance of different methods on VGG-16. As PCRR($\alpha $) gets smaller, the model size and FLOPs drop accordingly. Comparing non-CCM-loss and CCM-loss for VGG-16 trained from scratch, the result shows that incorporating CCM-loss brings no accuracy decline. When $\alpha $ equals 0.9, our method brings +0.27\% accuracy over our baseline while the parameters and FLOPs drop to 4.81M and 163.76M, the pruned ratios are 67.9\% and 47.9\%, respectively. When $\alpha $ equals 0.8, our method acquires an accuracy at 94.23\%, just a little drop -0.06\% from the baseline. Meanwhile, left 3.13M parameters and 110.99M FLOPs, pruned ratio are 79.1\% and 64.7\% respectively. When $\alpha $ equals 0.7, our method is overwhelmingly better than all the methods targeting high accuracy, shown in the first and second blocks of Tab.1, in all evaluation protocols. Accuracy at 94.12\% better than the highest 93.86\% from CHIP\cite{sui2021chip} for +0.26\% with 2.09M parameters and 75.69M FLOPs, the pruned ratios are 86\% and 75.9\%. When $\alpha $ equals 0.6, our methods retain only 1.40M parameters and 49.60M FLOPs from our baseline, pruned ratios are 90.6\% and 84.2\%. Our method brings another big step over HRank`s and CHIP`s targeting high model size and FLOPs reductions mode, shown in the third block of Tab.1. Still, accuracy at 93.64\% better than 93.18\% from CHIP for +0.46\%.

\begin{figure}[htbp]
\begin{center}
   \includegraphics[width=0.9\linewidth]{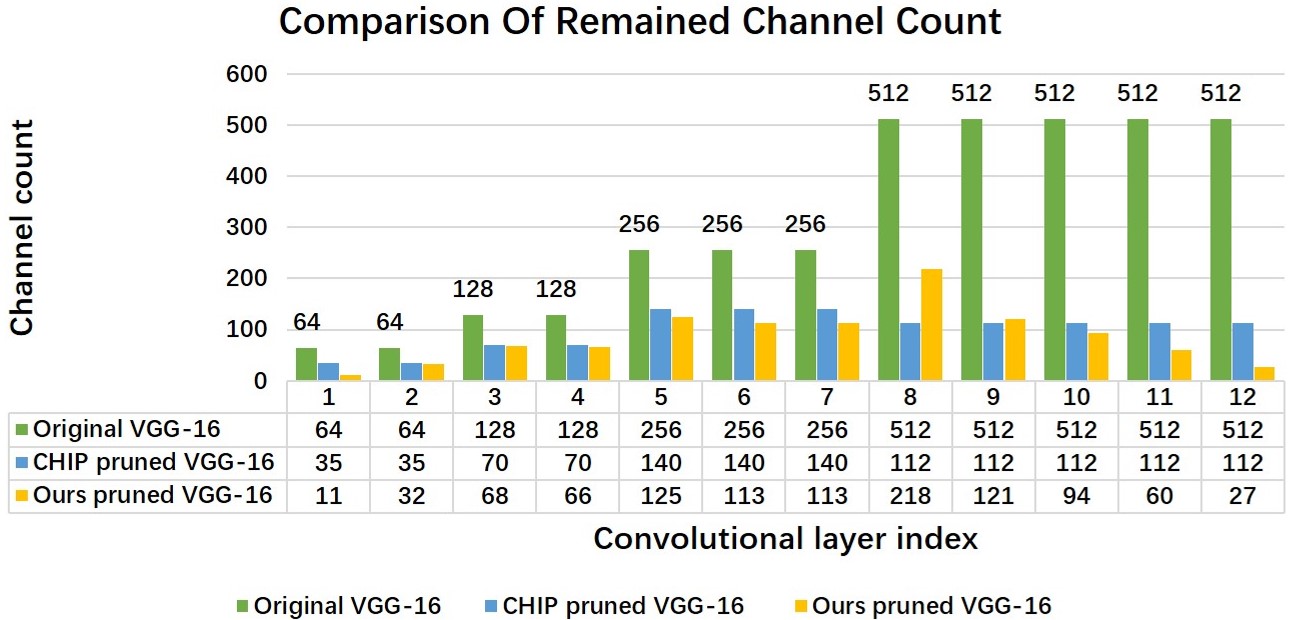}
\end{center}
   \caption{Comparison of remaining channel count of original, CHIP pruned and ours pruned VGG-16. Data came from $\star$ of Tab.1. Similar pruned network scale.}
\label{fig:long}
\end{figure}

Fig.5 shows the remained channel count of our method executed on VGG-16, we compare the result with CHIP. From Fig.5, we can get the following inference. Firstly, hard-coded compression for each layer can not guarantee the consistency and integrality of information flow through the network. Take the 8th layer as an example, the remained channels of our method are much more than CHIP, while our accuracy is better and the total parameter of the model is lesser. A Hard-code retention ratio for a layer that may need more parameters to get proper output may be harmful to the presentation ability of the model. On the other hand, taking the 12th layer as an example, our channel selection strategy dynamically set the retain ratio, leaving much lesser channels than CHIP. This means our method can exploit the maximum potential ability of CCM-loss to guarantee a minimum retain ratio. Secondly, comparing all through the 8th-12th layers, the remained channel of our method is declining. We suppose in recurring layer sequence, layers in the back of the sequence are more easily to be regulated by CCM-loss to generate LRR.

\subsubsection{ResNet-56 and ResNet-110 on Cifar-10}

\begin{table*}[htbp]
\footnotesize
\setlength{\abovecaptionskip}{0cm}
\begin{center}
\begin{tabular}{lccccc}

\toprule
\multirow{2}{*}{Method}  & \multicolumn{3}{c}{Top-1 Accuracy(\%)} & \multirow{2}{*} {Params($\downarrow$\%)} & \multirow{2}{*}{FLOPs($\downarrow$\%)} \\
\cmidrule{2-4}
 & Baseline & Pruned & $\Delta$  \\

\midrule
& & ResNet-56 & & &\\
$l_{1}$-norm\cite{li2016pruning} & 93.04 & 93.06 & +0.02 & 0.73M(13.7) & 90.90M(27.6)\\
NISP\cite{yu2018nisp}  & 93.04 & 93.01 & -0.03 & 0.49M(42.4) & 81.00M(35.5) \\
C-SGD-5/8\cite{ding2019centripetal}  & 93.39 & 93.31 & -0.08 & N/A & 48.94M(60.9) \\
GAL\cite{lin2019towards}  & 93.26 & 93.38 & +0.12 & 0.75M(11.8) & 78.30M(37.6) \\
Ghost-VGG-16 (s=2)\cite{han2020ghostnet}  & 93.00 & 92.70 & -0.30 & 0.43M(49.4) & 63.00M(49.6) \\
HRank\cite{lin2020hrank}  & 93.26 & 93.52 & +0.26 & 0.71M(16.8) & 88.72M(29.3) \\
CHIP\cite{sui2021chip}  & 93.26 & 94.16 & +0.90 & 0.48M(42.8) & 65.94M(47.4) \\
\hdashline
GAL\cite{lin2019towards}  & 93.26 & 91.58 & -1.68 & 0.29M(65.9) & 49.99M(60.2) \\
LASSO\cite{he2017channel}  & 92.80 & 91.80 & -1.00 & N/A & 62.00M(50.6) \\
HRank\cite{lin2020hrank}  & 93.26 & 90.72 & -2.54 & 0.27M(68.1) & 32.52M(74.1) \\
$\star$ CHIP\cite{sui2021chip}  & 93.26 & 92.05 & -1.21 & 0.24M(71.8) & 34.79M(72.3) \\
DepGraph\cite{fang2023depgraph}  & 93.53 & 93.77 & +0.24 & N/A & 60.48M(52.4) \\
\hdashline
Baseline(non-CCM-loss,unpruned)  & 94.52 & - & - & 0.85M(0.0) & 127.62M(0.0) \\
Baseline(CCM-loss,unpruned)  & 93.75 & - & - & - & - \\
Ours(PCRR=0.9)  & 93.75 & 94.00 & +0.25 & 0.54M(35.9) & 78.69M(38.3) \\
Ours(PCRR=0.8)  & 93.75 & 93.78 & +0.03 & 0.38M(55.9) & 55.63M(56.4) \\
Ours(PCRR=0.7)  & 93.75 & 92.74 & -1.01 & 0.25M(70.3) & 38.46M(69.9) \\
$\star$ Ours(PCRR=0.65)  & 93.75 & \textbf{92.40} & -1.35 & \textbf{0.20M(75.6)} & \textbf{32.10M(74.8)} \\
Ours(PCRR=0.6)  & 93.75 & 91.87 & -1.88 & 0.17M(80.2) & 26.56M(79.2) \\
\midrule
& & ResNet-110 & & &\\
$l_{1}$-norm\cite{li2016pruning} & 93.53 & 93.30 & -0.23 & 1.16M(32.4) & 155.00M(38.7)\\

HRank \cite{lin2020hrank} & 93.50 & 94.23 & +0.73 & 1.04M(39.4) & 148.70M(41.2) \\
CHIP\cite{sui2021chip}  & 93.50 & 94.44 & +0.94 & 0.89M(48.3) & 121.09M(52.1) \\

\hdashline
GAL\cite{lin2019towards}  & 93.50 & 92.74 & -0.76 & 0.95M(44.8) & 130.20M(48.5) \\
HRank\cite{lin2020hrank}  & 93.50 & 92.65 & -0.85 & 0.53M(68.7) & 79.30M(68.6) \\
CHIP\cite{sui2021chip}  & 93.50 & 93.63 & +0.13 & 0.54M(68.3) & 71.69M(71.6) \\
\hdashline
Baseline(non-CCM-loss,unpruned)  & 94.86 & - & - & 1.73M(0.0) & 257.08M(0.0) \\
Baseline(CCM-loss,unpruned)  & 94.14 & - & - & - & - \\
Ours(PCRR=0.9)  & 94.14 & \textbf{94.54} & +0.40 & 1.00M(41.9) & 136.71M(46.8) \\
Ours(PCRR=0.8)  & 94.14 & \textbf{94.19} & +0.05 & 0.67M(61.3) & 94.04M(63.4) \\
Ours(PCRR=0.75)  & 94.14 & \textbf{93.8} & -0.34 & 0.55M(68.4) & 78.59M(69.4) \\
Ours(PCRR=0.7)  & 94.14 & \textbf{93.35} & -0.79 & \textbf{0.45M(74.2)} & \textbf{64.73M(74.8)} \\
\bottomrule
\end{tabular}
\end{center}
\caption{ResNet-56 and ResNet-110 on Cifar-10.}
\label{fig:long}
\label{fig:onecol}
\end{table*}


\begin{table*}
\footnotesize
\setlength{\abovecaptionskip}{0cm}
\begin{center}
\begin{tabular}{lccccccccccccccccc}

\toprule
Channel index & 1 & 2 & $\cdot\cdot\cdot$ & 15 & 16 & 17 & $\cdot\cdot\cdot$ & 21 & 22 & 23& $\cdot\cdot\cdot$ & 29 & 30 & 31& $\cdot\cdot\cdot$ &54& 55\\
\midrule
Original & 16 & 16 & $\cdot\cdot\cdot$ & 16 & 16 & 16 & $\cdot\cdot\cdot$ & 32 & 32 & 32& $\cdot\cdot\cdot$ &32&32&32& $\cdot\cdot\cdot$ &64&64\\
CHIP\cite{sui2021chip} & 16 & 8 & $\cdot\cdot\cdot$ & 9 & 8 & 9 & $\cdot\cdot\cdot$ & 19 & 12 & 19& $\cdot\cdot\cdot$ &19&12&19& $\cdot\cdot\cdot$ &19&64\\
Our & 16 & 8 & $\cdot\cdot\cdot$ & 9 & \textbf{1} & 9 & $\cdot\cdot\cdot$ & 14 & \textbf{1} & 15& $\cdot\cdot\cdot$ &18&\textbf{1}&18& $\cdot\cdot\cdot$ &29&64 \\
\bottomrule
\end{tabular}
\end{center}
\caption{Retained channel comparison of ResNet-56. Data came from $\star$ of Tab.2.}
\label{fig:long}
\label{fig:onecol}
\end{table*}


\begin{table*}[htbp]
\footnotesize
\setlength{\abovecaptionskip}{0cm}
\begin{center}
\begin{tabular}{lcccccccc}
\toprule
\multirow{2}{*}{Method}  & \multicolumn{3}{c}{Top-1 Accuracy(\%)} & \multicolumn{3}{c}{Top-5 Accuracy(\%)} & \multirow{2}{*} {Params} & \multirow{2}{*}{FLOPs} \\
\cmidrule{2-7}
& Baseline & Pruned & $\Delta$ & Baseline & Pruned & $\Delta$ & $\downarrow$\% & $\downarrow$\%  \\

\midrule
ThiNet\cite{luo2017thinet} & 72.88 & 72.04 & -0.84 & 91.14 & 90.67 & -0.47 & 33.7 & 36.8 \\
SFP\cite{he2018soft} & 76.15 & 74.61 & -1.54 & 92.87 & 92.06 & -0.81 & N/A & 41.8 \\
FPGM\cite{he2019filter} & 76.15 & 75.59 & -0.56 & 92.87 & 92.63 & -0.24 & 37.5 & 42.2 \\
Taylor\cite{molchanov2019importance}  & 76.18 & 74.50 & -1.68 & N/A & N/A & N/A & 44.5 & 44.9 \\
C-SGD\cite{ding2019centripetal}  & 75.33 & 74.93 & -0.40 & 92.56 & 92.27 & -0.29 & N/A & 46.2 \\
GAL\cite{lin2019towards}  & 76.15 & 71.95 & -4.20 & 92.87 & 90.94 & -1.93 & 16.9 & 43 \\
RRBP\cite{zhou2019accelerate} & 76.10 & 73 & -3.10 & 92.90 & 91.00 & -1.90 & N/A & 54.5 \\
PFP\cite{liebenwein2019provable}  & 76.13 & 75.91 & -0.22 & 92.87 & 92.81 & -0.06 & 18.1 & 10.8 \\
Autopruner\cite{luo2020autopruner}  & 76.15 & 74.76 & -1.39 & 92.87 & 92.15 & -0.72 & N/A & 48.7 \\
HRank\cite{lin2020hrank}  & 76.15 & 74.98 & -1.17 & 92.87 & 92.33 & -0.54 & 36.6 & 43.7 \\
SCOP\cite{tang2020scientific}  & 76.15 & 75.95 & -0.20 & 92.87 & 92.79 & -0.08 & 42.8 & 45.3 \\
CHIP\cite{sui2021chip}  &76.15 & 76.30 & +0.15 & 92.87 & 93.02 & +0.15 & 40.8 & 44.8 \\
DepGraph\cite{fang2023depgraph} & 76.15 & 75.83 & -0.32 & N/A & N/A & N/A & N/A & 51.8 \\
\hdashline
Baseline(CCM-loss,unpruned)  & 76.46 & - & - & 92.96 &-&-& - & - \\
Ours(PCRR=0.9)  & 76.46 & \textbf{76.42} & -0.04 &92.96&93.04&+0.28& 24.5 & 28.5 \\
Ours(PCRR=0.8)  & 76.46 & \textbf{76.23} & -0.23 &92.96&92.87&+0.04& 42.8 & 47.3 \\
Ours(PCRR=0.75)  & 76.46 & \textbf{75.69} & -0.77 &92.96& 92.03&-0.93& \textbf{50.4} & \textbf{55.0} \\
Ours(PCRR=0.5) & 76.46 & 72.16 & -3.70 &92.96&91.03&-1.93& \textbf{78.5} & \textbf{79.9} \\

\bottomrule
\end{tabular}
\end{center}
\caption{ResNet-50 on ImageNet.}
\label{fig:long}
\label{fig:onecol}
\end{table*}

Tab.2 shows the performance of different methods on ResNet-56 and ResNet-110. For ResNet-56, when $\alpha $ equals 0.65, our method brings 92.40\% accuracy with 0.20M parameters and 32.10M FLOPs, and the pruned ratios are 75.6\% and 74.8\% respectively. Better than the best of targeting high model size and FLOPs reductions settings of all methods, shown in the 2nd block of Tab.2, in both accuracy and pruned ratio. For ResNet-110, when $\alpha $ equals 0.7, our method brings 93.35\% accuracy and the remained parameters and FLOPs are 0.45M and 64.73M, and the pruned ratios are 74.2\% and 64.73\%.

In Tab.3, we compare the remained channel count of our method executed on ResNet-56 with the result of CHIP\cite{sui2021chip}. Fascinatingly, there are three layers(16,22,30) in ResNet-56 pruned by our method leaving only one channel, meanwhile, the accuracy at 92.40\% is still maintained at a good level. In fact, from $\alpha $ equals 0.9, these layers already left only one channel. This phenomenon also appears in our implementation of ResNet-110. These results verified the ability of CCM-loss to produce more LRR and polarize channel importance. They also verified our channel selection strategy can dynamically select retain ratio for each layer and push the prune ratio to the limit. Another observation that should not be neglected is the indexes of the one-layer-left channels are not always the same in repeated train-from-scratch experiments, this may provide us with some intuition to understand the neural network itself.

\subsubsection{ResNet-50 On ImageNet}
Tab.4 shows the results of ResNet-50 trained on the ImageNet dataset. When $\alpha $ equals 0.8 our method prunes 42.8\% parameters from the baseline model and 47.3\% FLOPs, the accuracy is 76.48\%, +0.02 over the baseline accuracy. When $\alpha $ equals 0.75, our method prunes 50.4\% parameters and 55.0\% FLOPs, the accuracy at 75.82\%. And when $\alpha $ equals 0.5, we chop 78.5\% and 79.9\% of parameters and FLOPs, respectively, the accuracy of 72.16\% is still not bad.

\section{Conclusion And Discussion}

In this paper, we design a data-driven CCM-loss term base on the correlation coefficient between feature map channels to encourage neural networks to learn more LRR on the filter level during training. This loss term provides us with another universal transcendental mathematical tool besides $L_{*}$-norm-based loss terms to intentionally let the network learn more redundancy. By changing the operator CCM-loss, it can either enhance or weaken linear relations between different channels. We focus on the minus version of CCM-loss, and we suppose the plus version can be useful in other tasks, wherever we need to push the features linearly away from each other. Theoretically, this CCM-loss can be conducted on any repeat sub-network structured units which \textbf{simultaneously provide their output} to generate more LRR. For example, attention heads in Transformer. We also designed a matching channel selection strategy for the CCM-loss to exploit its maximum potential for filter-level structured pruning. Our strategy mainly focuses on keeping the consistency and integrality of the principal component of the information flow. It can dynamically, rather than empirically hard-coded, set the retain ratio for each layer, which can provide us with more accurate surgery for the model to acquire a higher compression ratio. By adjusting PCRR($\alpha$), we can control the final volume of the pruned model to fit application scenarios, memory or timing limitations. The three stages of our framework, generating more LRR, importance evaluation criteria, and pruning strategy, are separable structured processes. Future research can focus on any of them.
\bibliography{aaai24}

\begin{thebibliography}{30}
\providecommand{\natexlab}[1]{#1}

\bibitem[{Ding et~al.(2019)Ding, Ding, Guo, and Han}]{ding2019centripetal}
Ding, X.; Ding, G.; Guo, Y.; and Han, J. 2019.
\newblock Centripetal sgd for pruning very deep convolutional networks with
  complicated structure.
\newblock In \emph{Proceedings of the IEEE/CVF conference on computer vision
  and pattern recognition}, 4943--4953.

\bibitem[{Fang et~al.(2023)Fang, Ma, Song, Mi, and Wang}]{fang2023depgraph}
Fang, G.; Ma, X.; Song, M.; Mi, M.~B.; and Wang, X. 2023.
\newblock Depgraph: Towards any structural pruning.
\newblock In \emph{Proceedings of the IEEE/CVF Conference on Computer Vision
  and Pattern Recognition}, 16091--16101.

\bibitem[{Gholami et~al.(2022)Gholami, Kim, Dong, Yao, Mahoney, and
  Keutzer}]{gholami2022survey}
Gholami, A.; Kim, S.; Dong, Z.; Yao, Z.; Mahoney, M.~W.; and Keutzer, K. 2022.
\newblock A survey of quantization methods for efficient neural network
  inference.
\newblock In \emph{Low-Power Computer Vision}, 291--326. Chapman and Hall/CRC.

\bibitem[{Han et~al.(2020)Han, Wang, Tian, Guo, Xu, and Xu}]{han2020ghostnet}
Han, K.; Wang, Y.; Tian, Q.; Guo, J.; Xu, C.; and Xu, C. 2020.
\newblock Ghostnet: More features from cheap operations.
\newblock In \emph{Proceedings of the IEEE/CVF conference on computer vision
  and pattern recognition}, 1580--1589.

\bibitem[{He et~al.(2016)He, Zhang, Ren, and Sun}]{he2016deep}
He, K.; Zhang, X.; Ren, S.; and Sun, J. 2016.
\newblock Deep residual learning for image recognition.
\newblock In \emph{Proceedings of the IEEE conference on computer vision and
  pattern recognition}, 770--778.

\bibitem[{He et~al.(2018)He, Kang, Dong, Fu, and Yang}]{he2018soft}
He, Y.; Kang, G.; Dong, X.; Fu, Y.; and Yang, Y. 2018.
\newblock Soft filter pruning for accelerating deep convolutional neural
  networks.
\newblock In \emph{Proceedings of the 27th International Joint Conference on
  Artificial Intelligence}, 2234--2240.

\bibitem[{He et~al.(2019)He, Liu, Wang, Hu, and Yang}]{he2019filter}
He, Y.; Liu, P.; Wang, Z.; Hu, Z.; and Yang, Y. 2019.
\newblock Filter pruning via geometric median for deep convolutional neural
  networks acceleration.
\newblock In \emph{Proceedings of the IEEE/CVF conference on computer vision
  and pattern recognition}, 4340--4349.

\bibitem[{He, Zhang, and Sun(2017)}]{he2017channel}
He, Y.; Zhang, X.; and Sun, J. 2017.
\newblock Channel pruning for accelerating very deep neural networks.
\newblock In \emph{Proceedings of the IEEE international conference on computer
  vision(ICCV)}, 1389--1397.

\bibitem[{Hu et~al.(2022)Hu, Li, Liu, Chen, Wang, and Liu}]{hu2022teacher}
Hu, C.; Li, X.; Liu, D.; Chen, X.; Wang, J.; and Liu, X. 2022.
\newblock Teacher-Student Architecture for Knowledge Learning: A Survey.
\newblock \emph{arXiv preprint arXiv:2210.17332}.

\bibitem[{Hu et~al.(2016)Hu, Peng, Tai, and Tang}]{hu2016network}
Hu, H.; Peng, R.; Tai, Y.-W.; and Tang, C.-K. 2016.
\newblock Network trimming: A data-driven neuron pruning approach towards
  efficient deep architectures.
\newblock \emph{arXiv preprint arXiv:1607.03250}.

\bibitem[{Huang and Wang(2018)}]{huang2018data}
Huang, Z.; and Wang, N. 2018.
\newblock Data-driven sparse structure selection for deep neural networks.
\newblock In \emph{Proceedings of the European conference on computer vision
  (ECCV)}, 304--320.

\bibitem[{Krizhevsky, Hinton et~al.(2009)}]{krizhevsky2009learning}
Krizhevsky, A.; Hinton, G.; et~al. 2009.
\newblock Learning multiple layers of features from tiny images.

\bibitem[{Kumar et~al.(2021)Kumar, Shaikh, Li, Bilal, and
  Yin}]{kumar2021pruning}
Kumar, A.; Shaikh, A.~M.; Li, Y.; Bilal, H.; and Yin, B. 2021.
\newblock Pruning filters with L1-norm and capped L1-norm for CNN compression.
\newblock \emph{Applied Intelligence}, 51: 1152--1160.

\bibitem[{Li et~al.(2016)Li, Kadav, Durdanovic, Samet, and
  Graf}]{li2016pruning}
Li, H.; Kadav, A.; Durdanovic, I.; Samet, H.; and Graf, H.~P. 2016.
\newblock Pruning filters for efficient convnets.
\newblock \emph{arXiv preprint arXiv:1608.08710}.

\bibitem[{Liebenwein et~al.(2019)Liebenwein, Baykal, Lang, Feldman, and
  Rus}]{liebenwein2019provable}
Liebenwein, L.; Baykal, C.; Lang, H.; Feldman, D.; and Rus, D. 2019.
\newblock Provable filter pruning for efficient neural networks.
\newblock \emph{arXiv preprint arXiv:1911.07412}.

\bibitem[{Lin et~al.(2020)Lin, Ji, Wang, Zhang, Zhang, Tian, and
  Shao}]{lin2020hrank}
Lin, M.; Ji, R.; Wang, Y.; Zhang, Y.; Zhang, B.; Tian, Y.; and Shao, L. 2020.
\newblock Hrank: Filter pruning using high-rank feature map.
\newblock In \emph{Proceedings of the IEEE/CVF conference on computer vision
  and pattern recognition(CVPR)}, 1529--1538.

\bibitem[{Lin et~al.(2019)Lin, Ji, Yan, Zhang, Cao, Ye, Huang, and
  Doermann}]{lin2019towards}
Lin, S.; Ji, R.; Yan, C.; Zhang, B.; Cao, L.; Ye, Q.; Huang, F.; and Doermann,
  D. 2019.
\newblock Towards optimal structured cnn pruning via generative adversarial
  learning.
\newblock In \emph{Proceedings of the IEEE/CVF conference on computer vision
  and pattern recognition(CVPR)}, 2790--2799.

\bibitem[{Luo and Wu(2020)}]{luo2020autopruner}
Luo, J.-H.; and Wu, J. 2020.
\newblock Autopruner: An end-to-end trainable filter pruning method for
  efficient deep model inference.
\newblock \emph{Pattern Recognition}, 107: 107461.

\bibitem[{Luo, Wu, and Lin(2017)}]{luo2017thinet}
Luo, J.-H.; Wu, J.; and Lin, W. 2017.
\newblock Thinet: A filter level pruning method for deep neural network
  compression.
\newblock In \emph{Proceedings of the IEEE international conference on computer
  vision}, 5058--5066.

\bibitem[{Mi, Feng, and Huang(2022)}]{mi2022designing}
Mi, J.-X.; Feng, J.; and Huang, K.-Y. 2022.
\newblock Designing efficient convolutional neural network structure: A survey.
\newblock \emph{Neurocomputing}, 489: 139--156.

\bibitem[{Molchanov et~al.(2019)Molchanov, Mallya, Tyree, Frosio, and
  Kautz}]{molchanov2019importance}
Molchanov, P.; Mallya, A.; Tyree, S.; Frosio, I.; and Kautz, J. 2019.
\newblock Importance estimation for neural network pruning.
\newblock In \emph{Proceedings of the IEEE/CVF conference on computer vision
  and pattern recognition}, 11264--11272.

\bibitem[{Paszke et~al.(2017)Paszke, Gross, Chintala, Chanan, Yang, DeVito,
  Lin, Desmaison, Antiga, and Lerer}]{paszke2017automatic}
Paszke, A.; Gross, S.; Chintala, S.; Chanan, G.; Yang, E.; DeVito, Z.; Lin, Z.;
  Desmaison, A.; Antiga, L.; and Lerer, A. 2017.
\newblock Automatic differentiation in pytorch.

\bibitem[{Simonyan and Zisserman(2014)}]{simonyan2014very}
Simonyan, K.; and Zisserman, A. 2014.
\newblock Very deep convolutional networks for large-scale image recognition.
\newblock \emph{arXiv preprint arXiv:1409.1556}.

\bibitem[{Sui et~al.(2021)Sui, Yin, Xie, Phan, Aliari~Zonouz, and
  Yuan}]{sui2021chip}
Sui, Y.; Yin, M.; Xie, Y.; Phan, H.; Aliari~Zonouz, S.; and Yuan, B. 2021.
\newblock Chip: Channel independence-based pruning for compact neural networks.
\newblock \emph{Advances in Neural Information Processing Systems(NeurIPS)},
  34: 24604--24616.

\bibitem[{Tang et~al.(2020)Tang, Wang, Xu, Tao, Xu, Xu, and
  SCOP}]{tang2020scientific}
Tang, Y.; Wang, Y.; Xu, Y.; Tao, D.; Xu, C.; Xu, C.; and SCOP, C.~X. 2020.
\newblock Scientific control for reliable neural network pruning.
\newblock \emph{Neural Information Processing Systems (NeurIPS)}, 1(2): 7.

\bibitem[{Tropp and Webber(2023)}]{tropp2023randomized}
Tropp, J.~A.; and Webber, R.~J. 2023.
\newblock Randomized algorithms for low-rank matrix approximation: Design,
  analysis, and applications.
\newblock \emph{arXiv preprint arXiv:2306.12418}.

\bibitem[{Wen et~al.(2016)Wen, Wu, Wang, Chen, and Li}]{wen2016learning}
Wen, W.; Wu, C.; Wang, Y.; Chen, Y.; and Li, H. 2016.
\newblock Learning structured sparsity in deep neural networks.
\newblock \emph{Advances in neural information processing systems(NeurIPS)},
  29.

\bibitem[{Yu et~al.(2018)Yu, Li, Chen, Lai, Morariu, Han, Gao, Lin, and
  Davis}]{yu2018nisp}
Yu, R.; Li, A.; Chen, C.-F.; Lai, J.-H.; Morariu, V.~I.; Han, X.; Gao, M.; Lin,
  C.-Y.; and Davis, L.~S. 2018.
\newblock Nisp: Pruning networks using neuron importance score propagation.
\newblock In \emph{Proceedings of the IEEE conference on computer vision and
  pattern recognition(CVPR)}, 9194--9203.

\bibitem[{Yuan and Lin(2006)}]{yuan2006model}
Yuan, M.; and Lin, Y. 2006.
\newblock Model selection and estimation in regression with grouped variables.
\newblock \emph{Journal of the Royal Statistical Society: Series B (Statistical
  Methodology)}, 68(1): 49--67.

\bibitem[{Zhou et~al.(2019)Zhou, Zhang, Wang, and Tian}]{zhou2019accelerate}
Zhou, Y.; Zhang, Y.; Wang, Y.; and Tian, Q. 2019.
\newblock Accelerate cnn via recursive bayesian pruning.
\newblock In \emph{Proceedings of the IEEE/CVF International Conference on
  Computer Vision}, 3306--3315.

\end{thebibliography}

\end{document}